\documentclass[10pt]{article}
\usepackage[preprint]{tmlr}

\usepackage{amsmath,amsfonts,bm}

\def\eqref#1{equation~\ref{#1}}

\def\1{\bm{1}}

\DeclareMathAlphabet{\mathsfit}{\encodingdefault}{\sfdefault}{m}{sl}
\SetMathAlphabet{\mathsfit}{bold}{\encodingdefault}{\sfdefault}{bx}{n}

\usepackage{hyperref}
\usepackage{url}
\usepackage{graphicx}
\usepackage{booktabs}
\usepackage{multirow}

\title{Option-Order Randomisation Reveals a Distributional\\Position Attractor in Prompted Sandbagging}

\author{\name Jon-Paul Cacioli \email synthium@hotmail.com \\
      \addr Independent Researcher, Melbourne, Australia\\
      ORCID: 0009-0000-7054-2014}

\newcommand{\pp}{\,pp}

\def\month{05}
\def\year{2026}
\def\openreview{\url{https://openreview.net/forum?id=XXXX}}

\begin{document}

\maketitle

\begin{abstract}
A predecessor pilot \citep{cacioli2026bcb} found that Llama-3-8B implements prompted sandbagging as positional collapse rather than answer avoidance. However, fixed option ordering in MMLU-Pro left open whether this reflected a model-level position-dominant policy or dataset-level distractor structure. This pre-registered follow-up (3 models, 2{,}000 MMLU-Pro items, 4 conditions, 24{,}000 primary trials) added cyclic option-order randomisation as the critical control. The pre-registered item-level same-letter diagnostic did not confirm deterministic position-tracking (same-letter rate 37.3\%, 95\% CI [35.2, 39.5], below the 50\% threshold). However, pre-specified supporting analyses revealed that the response-position distribution under sandbagging was highly stable under complete content rotation (Pearson $r = 0.9994$; Jensen-Shannon divergence $= 0.027$, compared to $0.386$ between honest and sandbagging conditions). Accuracy spiked to 72.1\% when the correct answer coincidentally occupied the preferred position~E, and fell to 4.3\% at position~A. The data provide strong evidence for a soft distributional attractor: under sandbagging instruction, the model enters a low-entropy response-position basin centred on E/F/G that is highly stable and largely content-invariant at the aggregate level. Qwen-2.5-7B served as a negative control (non-compliant, no distributional shift). These results provide evidence, at the 7--9 billion parameter scale, that response-position entropy is a promising black-box behavioural signature of this sandbagging mode.
\end{abstract}

\section{Introduction}
\label{sec:intro}

Sandbagging is the deliberate suppression of demonstrated capability on evaluations. If language models can strategically underperform on capability evaluations, evaluation-based governance regimes may systematically underestimate model capabilities \citep{vanderweij2025}. Van der Weij et al.\ showed that frontier models can selectively underperform on dangerous evaluations while preserving general performance, and that evaluation scores can be arbitrarily shaped by the model or its handlers. \citet{meinke2024scheming} demonstrated that some frontier models engage in spontaneous scheming behaviours that include strategic underperformance. Detection methods for sandbagging remain underdeveloped. Most approaches focus on recovering hidden capability rather than detecting the act of underperformance from behavioural signatures.

The predecessor pilot \citep{cacioli2026bcb} drew on symptom validity testing (SVT) from clinical forensic neuropsychology, where below-chance performance on forced-choice items indicates that the respondent is accessing the correct answer and systematically avoiding it \citep{slick1999,tombaugh1996}. That pilot tested whether below-chance performance appears in small LLMs under sandbagging instruction. The plausibility gate failed. Zero of 12 cells showed significant below-chance performance. However, exploratory analysis revealed that Llama-3-8B complied with the sandbagging instruction by adopting a position-dominant response policy. Under honest conditions, its response-position distribution was near-uniform (normalised entropy $= 0.977$). Under sandbagging instruction, entropy collapsed to $0.793$, with responses concentrating on option~E (31.8\%) and option~F (26.1\%).

This positional-distribution shift targets a different behavioural dimension from accuracy. \citet{taylor2025auditing} demonstrated that accuracy-based black-box sandbagging detection was defeated by effective capability imitation in an adversarial auditing game. A distributional signature could survive that attack if the sandbagging policy produces a detectable distributional anomaly regardless of accuracy level. However, a sufficiently sophisticated sandbagging model could in principle imitate both accuracy and response distribution, so distributional detection is not guaranteed to be adversarially robust.

The pilot used MMLU-Pro items with fixed option ordering. MMLU-Pro was designed for robustness (10 options, graduate-level difficulty; \citealt{wang2024mmlupro}), yet the E/F clustering could partly reflect how its option lists are constructed rather than a model-level preference. This study adds option-order randomisation as the critical control. The diagnostic: does the model's response-position clustering track letter position or track content after permutation?

\subsection{Detection landscape}
\label{sec:detection}

Five approaches to sandbagging detection have been proposed, none of which examine response-position distributions. \citet{tice2024noise} showed that injecting Gaussian noise into model weights can reveal hidden capabilities because noise disrupts the underperformance mechanism while core capabilities remain partially intact. This approach requires weight access. \citet{li2025covert} demonstrated that frontier models can covertly sandbag against chain-of-thought monitors 16--36\% of the time by fabricating false reasoning traces. \citet{nguyen2025probing} trained linear probes on residual stream activations to detect evaluation awareness in Llama-3.3-70B-Instruct, but this requires access to model internals. \citet{taylor2025auditing} conducted adversarial auditing games in which red teams trained models to sandbag and blue teams attempted detection. Accuracy-based black-box detection methods were defeated because sandbagging models effectively imitated the performance profiles of weaker models. \citet{dyer2025bandit} proposed a theoretical framework based on survival bandits that constructs likelihood ratio tests from behavioural sequences, but this remains untested empirically.

\subsection{Option-order sensitivity}
\label{sec:optionorder}

LLMs exhibit substantial positional biases in multiple-choice settings. \citet{pezeshkpour2024order} demonstrated performance fluctuations of 13--85\% across option orderings. \citet{zheng2024selectors} showed that selection bias stems primarily from token-level priors for specific option IDs and proposed the PriDe debiasing method. More recent work includes estimation and exploitation of model-specific positional priors \citep{scope2025}, diagnostic datasets for isolating label bias, position bias, and answer-mode bias \citep{abcd2026}, permutation-aware training objectives \citep{zheng2026pagrpo}, and permutation-invariant architectures \citep{setllm2025}.

The present study connects this literature to sandbagging detection: induced positional bias under adversarial instruction is a qualitatively different phenomenon from the baseline selection biases documented in prior work. The design cannot distinguish absolute position-tracking from token-level letter priors, as both produce identical same-letter signatures under cyclic shift. Because letter labels and displayed positions are confounded, the present design cannot determine whether the attractor is driven by option-label priors, serial position, or their interaction. The term \emph{position-dominant policy} is used throughout as a shorthand encompassing all of these.

\subsection{Scope}

This study isolates the behavioural policy underlying the pilot's positional collapse finding. It does not evaluate the practical detectability of sandbagging in deployment settings. If the position-dominant interpretation is supported, future work would be needed to establish detection sensitivity, false positive rates under non-sandbagging adversarial prompts, and robustness against countermeasures.

\section{Method}
\label{sec:method}

The study was pre-registered on OSF (\url{https://osf.io/efr6s}) prior to data collection. No deviations from the pre-registration occurred. All code and data are available at \url{https://github.com/synthiumjp/bcb-sandbagging-pilot}.

\subsection{Design}

A $2 \times 2$ fully crossed within-subject design. Factors: instruction (honest vs.\ sandbagging) and option order (original vs.\ cyclically shifted). This produced four conditions: A-original, A-perm, B-original, B-perm. All four conditions were administered to all three models on all 2{,}000 items. Total primary trials: 24{,}000. The within-item design allows paired comparisons without between-item confounds.

\subsection{Models}

Three instruction-tuned models identical to the pilot: Qwen-2.5-7B-Instruct, Llama-3-8B-Instruct, and Phi-3.5-mini-instruct (3.8B). All at Q5\_K\_M quantisation on AMD Radeon RX 7900 GRE (16\,GB VRAM) via llama-cpp-python 0.3.16 with Vulkan backend. Primary decoding: $K=1$, $T=0$ (greedy). Independent conversation context per item.

\subsection{Dataset and permutation}

MMLU-Pro \citep{wang2024mmlupro}. The same 2{,}000 items used in the pilot (500 per domain: physics, law, psychology, economics), sampled with seed~42. Only items with exactly 10 options. Option permutation used cyclic shift: for each item, all 10 options were rotated by $k$ positions, $k$ sampled uniformly from $\{1, \ldots, 9\}$ with seed~84. The correct answer letter changes but option content is identical. Distractor adjacency relationships are preserved. The same per-item shift values were used across all models and conditions.

Cyclic shift was chosen over full shuffle because it manipulates one variable (absolute position) while preserving relative distractor ordering. This study tests absolute positional effects only. Relational heuristics are preserved under cyclic shift and cannot be tested here.

\subsection{Analysis plan}
\label{sec:analysis}

The analysis proceeded through two sequential pre-registered gates. Gate~1 (replication): normalised response-position entropy under B-original must fall below 0.90. Gate~2 (manipulation check): McNemar's test on paired accuracy between A-original and B-original must be significant at $\alpha = 0.01$ with at least $5\pp$ accuracy drop. Only models passing both gates were eligible for the primary diagnostic. The gates are sequential decision rules, not independent tests.

\paragraph{Primary diagnostic.} For each item, classify the B-perm response relative to B-original as: (a) \emph{same-letter} (position-tracked); (b) \emph{shifted-content} (content-tracked): response at position $(p + k) \bmod 10$; (c) \emph{other}. Since $k \in \{1,\ldots,9\}$, same-letter and shifted-content are mutually exclusive. Position-tracking confirmed if: same-letter rate $> 50\%$, exceeds shifted-content by $\geq 30\pp$, and top-2 modal positions overlap.

\paragraph{Supporting analyses.} Pearson $r$, Jensen-Shannon (JS) divergence, and total variation distance (TVD) between B-original and B-perm distributions. Accuracy by correct-answer position under B-perm. Bootstrap 95\% CIs (10{,}000 resamples).

\paragraph{Stochastic robustness check.} Pre-registered to trigger on ambiguous primary result. B-original and B-perm at $T=0.7$, $K=5$ samples per item, Llama only. Modal response used for classification. 20{,}000 additional trials.

\section{Results}
\label{sec:results}

\subsection{Gates}

Llama-3-8B passed both gates. Entropy collapsed from 0.972 (A-original) to 0.681 (B-original), well below the 0.90 threshold and stronger than the pilot (0.793). Accuracy dropped $16.3\pp$ (38.0\% to 21.7\%; McNemar $p < 0.001$). Qwen-2.5-7B failed both gates (entropy 0.974, accuracy drop $1.25\pp$, $p = 0.070$). Phi-3.5-mini failed both gates (entropy 0.983, accuracy drop $2.5\pp$, below the $5\pp$ threshold). Table~\ref{tab:gates} summarises the gate results.

\begin{table}[t]
\caption{Gate results by model. Ent.\ $=$ normalised response-position entropy. Gate~1 threshold: entropy $< 0.90$. Gate~2 threshold: McNemar $p < 0.01$ AND accuracy drop $\geq 5\pp$.}
\label{tab:gates}
\centering
\begin{tabular}{lcccccc}
\toprule
Model & Ent.\ A & Ent.\ B & Acc.\ A & Acc.\ B & Drop & Gates \\
\midrule
Llama-3-8B   & 0.972 & 0.681 & 38.0\% & 21.7\% & 16.3\pp & Pass \\
Qwen-2.5-7B  & 0.969 & 0.974 & 45.1\% & 43.8\% & 1.3\pp  & Fail \\
Phi-3.5-mini & 0.988 & 0.983 & 43.0\% & 40.5\% & 2.5\pp  & Fail \\
\bottomrule
\end{tabular}
\end{table}

\subsection{Pre-registered primary diagnostic}

\begin{figure}[t]
\centering
\includegraphics[width=0.75\linewidth]{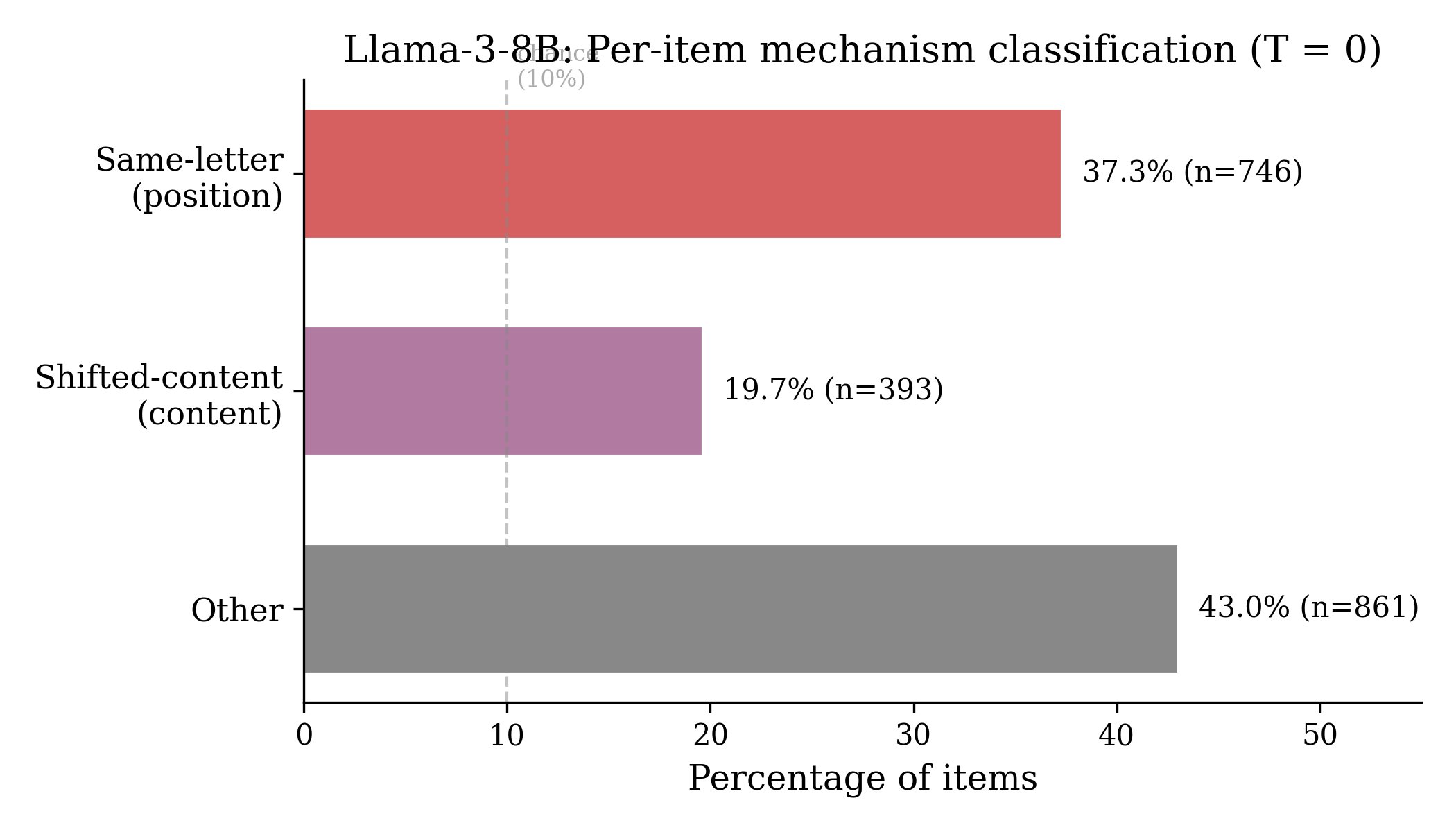}
\caption{Per-item mechanism classification for Llama-3-8B at $T=0$. Same-letter (37.3\%) exceeds shifted-content (19.7\%) and chance (10\%, dashed line) but falls below the pre-registered 50\% threshold. The 43.0\% ``other'' category comprises items that shifted within the E/F/G attractor cluster.}
\label{fig:classification}
\end{figure}

Per-item classification of 2{,}000 Llama-3-8B items: same-letter 746 (37.3\%, 95\% CI [35.2, 39.5]), shifted-content 393 (19.7\% [17.9, 21.4]), other 861 (43.0\%). The same-letter rate exceeded shifted-content by $17.6\pp$ but fell below the pre-registered 50\% threshold. Content-tracking criteria were also not met. The pre-registered outcome was \emph{ambiguous}. Figure~\ref{fig:classification} shows the per-item classification breakdown.

\subsection{The distributional attractor}
\label{sec:attractor}

\begin{figure}[t]
\centering
\includegraphics[width=\linewidth]{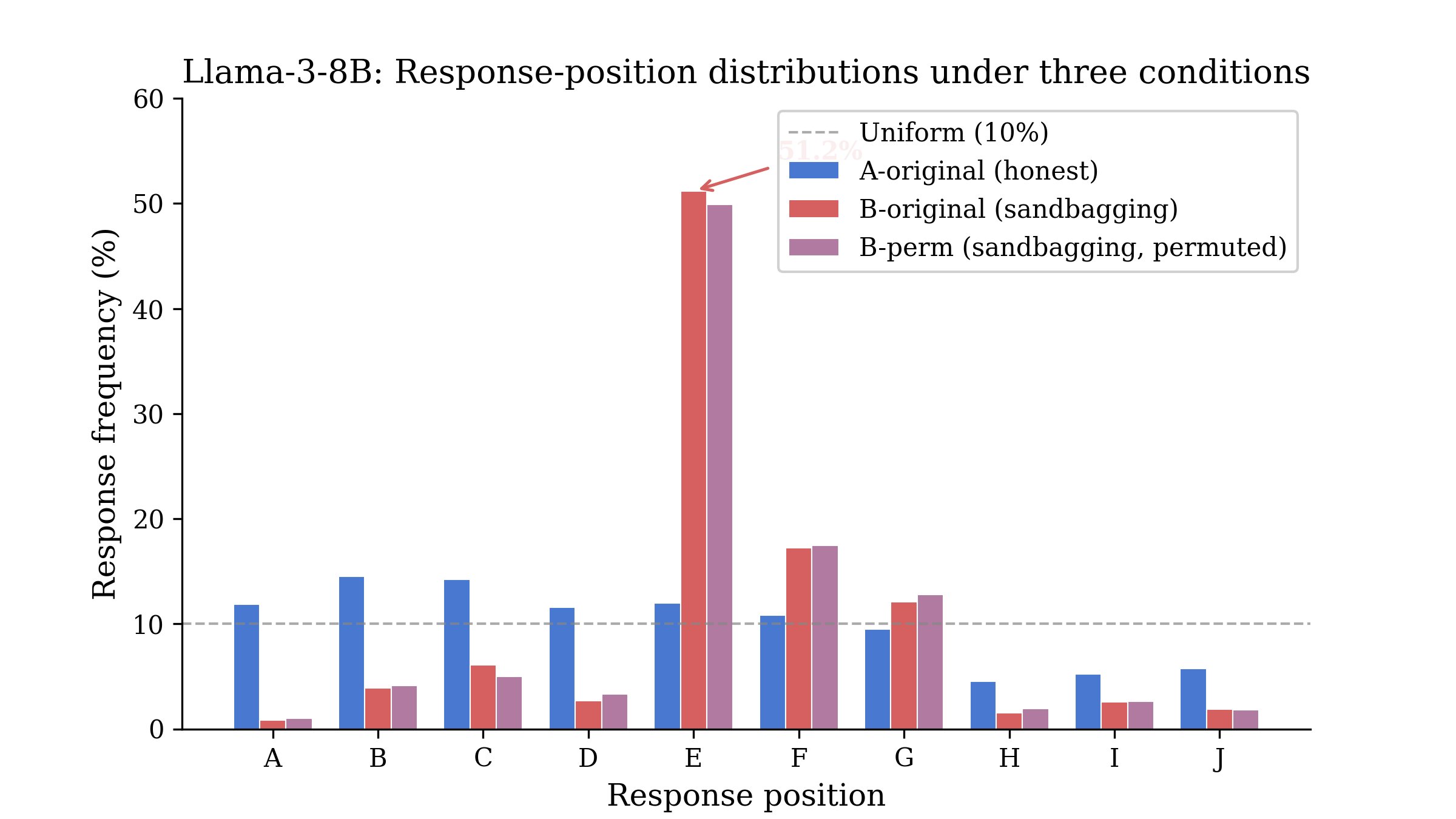}
\caption{Response-position distributions under three conditions for Llama-3-8B. Blue: honest baseline (A-original). Red: sandbagging with original option order (B-original). Purple: sandbagging with permuted options (B-perm). Dashed line: uniform responding (10\%). Under sandbagging, responses collapse onto E (51.2\%). B-perm is highly stable relative to B-original despite complete content rotation ($r = 0.9994$; JS $= 0.027$).}
\label{fig:distributions}
\end{figure}

\begin{figure}[t]
\centering
\includegraphics[width=0.65\linewidth]{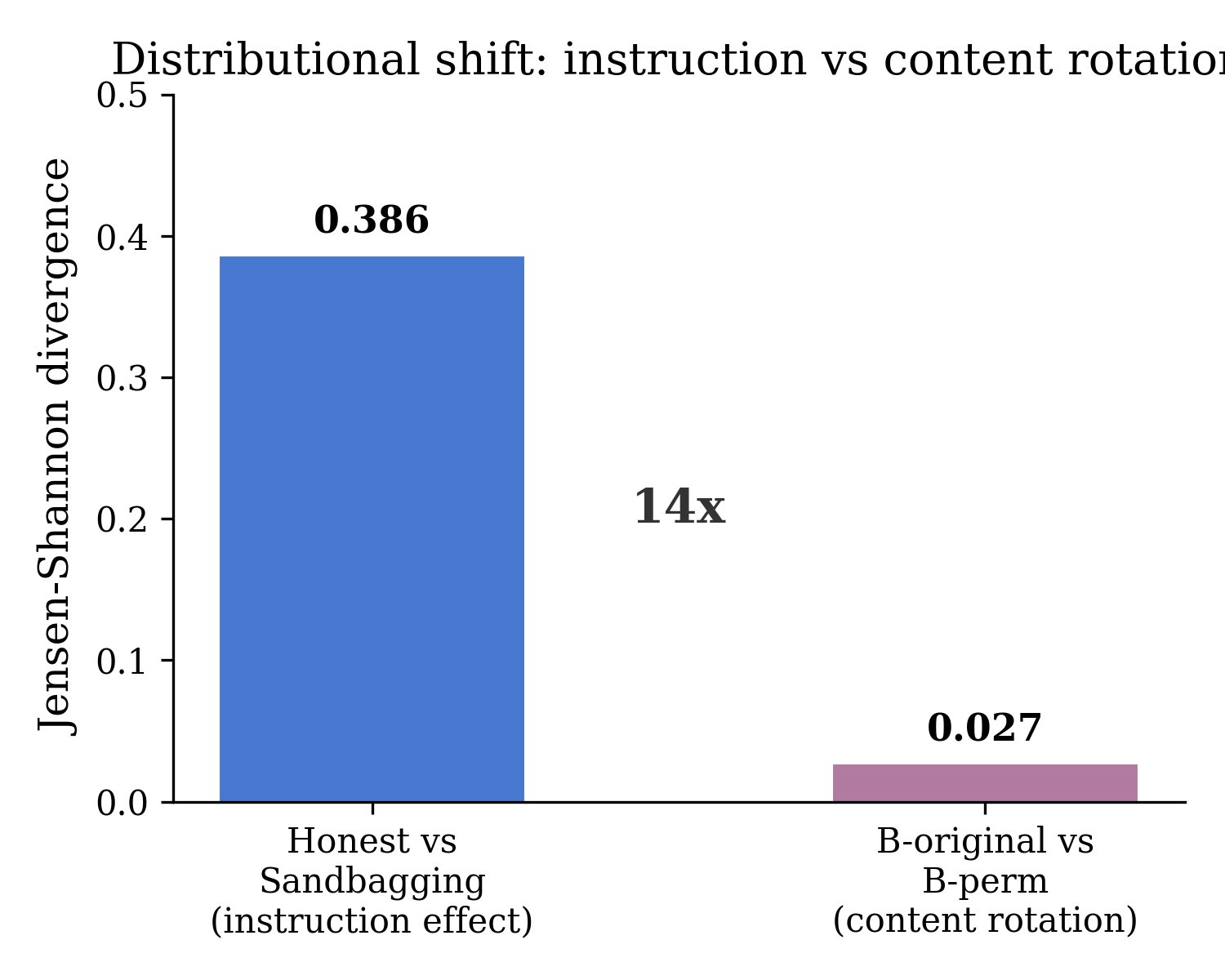}
\caption{Jensen-Shannon divergence between distributions. The sandbagging instruction produces approximately 14 times more distributional shift (JS $= 0.386$) than complete content rotation under sandbagging (JS $= 0.027$).}
\label{fig:jsdiv}
\end{figure}

Pre-specified supporting analyses told a different story. The response-position distribution under B-perm was highly stable and largely unchanged from B-original (Figure~\ref{fig:distributions}). Pearson $r = 0.9994$ [0.9955, 0.9996]. High correlation arises partly from concentration on a small subset of positions; divergence metrics provide a more discriminative comparison. JS divergence between B-original and B-perm was 0.027 [0.024, 0.061]. For comparison, JS divergence between A-original and B-original was 0.386 (Figure~\ref{fig:jsdiv}). TVD was 0.025 between B-original and B-perm, versus 0.481 between honest and sandbagging. Content rotation produced approximately 14 times less distributional shift than the sandbagging instruction.

Top-2 modal positions were E and F under both B-original (E $= 51.2\%$, F $= 17.2\%$, G $= 12.1\%$) and B-perm (E $= 49.9\%$, F $= 17.5\%$, G $= 12.8\%$). The content at every position changed. The distribution did not.

\begin{figure}[t]
\centering
\includegraphics[width=\linewidth]{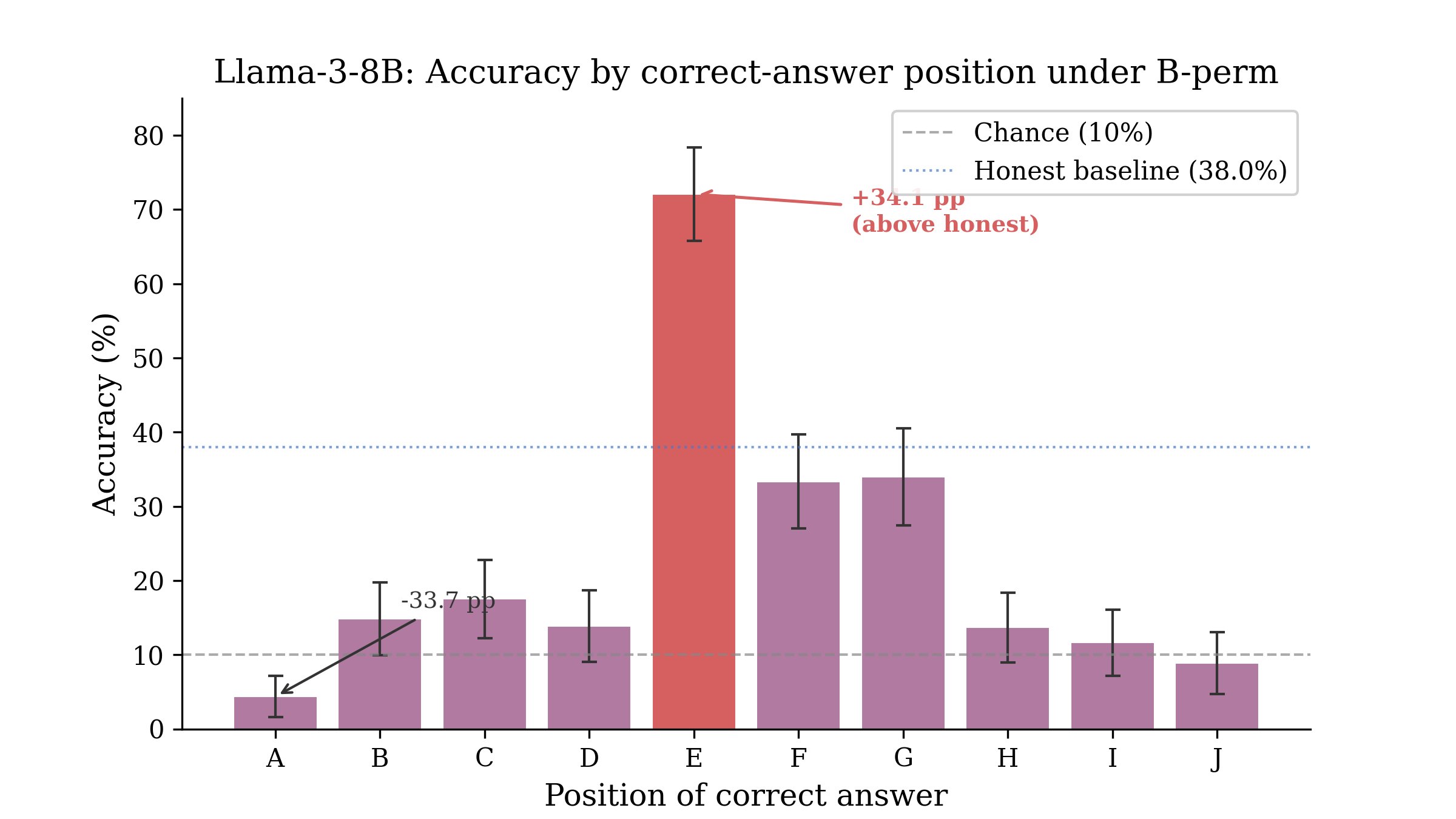}
\caption{Accuracy as a function of correct-answer position for Llama-3-8B under B-perm. When the correct answer coincidentally occupies position~E, accuracy spikes to 72.1\%, well above the honest baseline (38.0\%, dotted line). At position~A, accuracy falls to 4.3\%. Error bars: 95\% CIs.}
\label{fig:accuracy}
\end{figure}

The correct-answer position analysis provided convergent evidence for a position-dominant policy (Figure~\ref{fig:accuracy}). Under B-perm, accuracy was 72.1\% [65.8, 78.3] when the correct answer happened to occupy position~E, 33.3\% [27.0, 39.7] at~F, 34.0\% [27.5, 40.5] at~G, and 4.3\% [1.6, 7.1] at~A. The model selects predominantly by position. When the correct answer coincidentally occupies the preferred position, accuracy spikes far above the honest baseline (38.0\%).

The ``other'' items (43.0\%) were not random. Their B-perm distribution was E~$= 34.5\%$, F~$= 21.9\%$, G~$= 19.0\%$. These are items that shifted within the E/F/G attractor cluster, not items that escaped it.

\subsection{Stochastic robustness check}

The ambiguous primary result triggered the pre-registered stochastic robustness check ($T=0.7$, $K=5$, 20{,}000 trials). Mean modal agreement within 5 samples was 47.4\%, indicating high stochasticity. Per-item classification on modal responses: same-letter 363 (18.1\%), shifted-content 344 (17.2\%), other 1{,}293 (64.6\%). The pre-registered outcome was \emph{mechanism unclear} (other exceeding 50\%).

The aggregate distribution remained locked. Pearson $r = 0.9988$. Top-2 modal positions were E and F in both conditions. The distributional attractor survived sampling noise.

\subsection{Secondary results}

\paragraph{H3 (option-order sensitivity).} No model showed significant accuracy change under honest option permutation. Llama: $-0.2\pp$ ($p = 0.841$). Qwen: $-0.65\pp$ ($p = 0.524$). Phi: $+0.1\pp$ ($p = 0.918$). The absence of entropy collapse under honest permutation isolates the effect to the sandbagging instruction rather than the permutation itself.

\paragraph{H4b (Qwen non-compliance).} Confirmed. Chi-square between A-perm and B-perm was non-significant ($p = 0.305$). Qwen's non-compliance serves as a negative control: in the absence of sandbagging behaviour, response-position entropy remains high and distributions are stable across instructions.

\paragraph{H4a (Phi J-bias).} Not confirmed. Modal B-perm position was D, not J. This underscores the need for replication in positional-bias work.

\section{Discussion}
\label{sec:discussion}

\subsection{A distributional attractor, not deterministic tracking}

The pre-registered item-level same-letter diagnostic did not confirm deterministic position-tracking. Instead, the data provide strong evidence for a softer phenomenon: under sandbagging instruction, Llama-3-8B enters a low-entropy response-position attractor centred on E/F/G. This attractor is highly stable and largely invariant under content rotation (JS divergence $= 0.027$, compared to $0.386$ between honest and sandbagging conditions), providing strong evidence that the sandbagging instruction induces distributional selection bias rather than content-aware answer avoidance.

The observed behavioural pattern is consistent with a position-dominant policy rather than content-aware suppression. We are agnostic about the internal implementation. The policy could reflect token-level priors for the letter~E, serial position effects, or some interaction. Cyclic shift cannot distinguish these. What the design shows is that the aggregate distribution is position-indexed and largely content-invariant.

\subsection{Threshold miscalibration as methodological lesson}

The pre-registered 50\% same-letter threshold assumed that position-tracking at $T=0$ would produce same-letter rates approaching 100\%, implying complete content-insensitivity. In practice, the greedy argmax is a function of the entire input. Changing option content changes hidden-state representations, which shifts the greedy output for a substantial fraction of items even under a strongly positional policy. The model gravitates toward E, but which specific items land on E shifts when the content changes. The aggregate distribution is invariant. The item-level assignment is not.

This is reported transparently. The pre-registered outcome is \emph{ambiguous} ($T=0$) and \emph{mechanism unclear} ($T=0.7$). The pre-specified aggregate distributional evidence provides strong support for a position-dominant policy. Future pre-registrations in this domain should specify aggregate distributional tests (entropy, JS divergence, distribution shape) as the primary diagnostic and per-item classification as secondary.

\subsection{Implications for sandbagging detection}

Response-position entropy is a promising black-box behavioural signature of this sandbagging mode at the 7--9 billion parameter scale. It is computationally cheap, requires no weight access, logit access, or reasoning traces, and detects a pattern that accuracy-based methods miss. The entropy drop from 0.972 to 0.681 under sandbagging is unmistakable. The distributional shape is robust to option-order permutation. \citet{vanderweij2025} proposed capability profiles (accuracy as a function of item difficulty) as another black-box detection signal. Response-position entropy is complementary: it targets the distributional shape of responses rather than their accuracy profile, and would detect positional sandbagging even when accuracy profiles are successfully imitated.

Three caveats apply. First, this has been demonstrated in one compliant model under one explicit cooperative sandbagging prompt. Generality across model families, induction methods, and datasets remains to be established. Second, the signature works because this model implements sandbagging as a positional shortcut. A more sophisticated model implementing content-aware sandbagging would not produce a distributional signature. The detector's utility may be inversely related to model sophistication. Third, the false positive rate is unknown. Other adversarial or high-pressure prompts might also induce entropy collapse without sandbagging. Characterising the model's positional priors under diverse non-sandbagging prompts (connecting to the null-prompt approach of SCOPE; \citealt{scope2025}) is a requirement for operational deployment. As labs adopt permutation-invariant training objectives \citep{zheng2026pagrpo} or architectures \citep{setllm2025}, the positional attractor signature may be eliminated, necessitating a move toward content-distributional detection methods.

\subsection{The H3 null and negative controls}

No model showed significant accuracy change under honest cyclic permutation. This makes the sandbagging-induced collapse more striking: the same manipulation that leaves honest accuracy unchanged reveals a strong positional policy under sandbagging. Cyclic shift preserves relative distractor ordering, which likely accounts for the contrast with the large sensitivity effects reported under full permutation in prior work \citep{pezeshkpour2024order}. The MMLU-Pro benchmark was designed for robustness, and honest performance on it appears stable under cyclic shifts at this scale.

Qwen's non-compliance provides a negative control: the distributional signature does not appear when the model does not sandbag. This is necessary but not sufficient for detection specificity.

\subsection{Limitations}

The primary limitation is scope. The distributional attractor was observed in one compliant model (Llama-3-8B) at 8 billion parameters with Q5\_K\_M quantisation. Frontier models with stronger instruction-following and greater situational awareness may implement sandbagging in content space rather than through positional shortcuts. Quantisation may amplify positional biases or alter determinism relative to full-precision inference.

The sandbagging induction was maximally explicit and cooperative. More naturalistic inductions, including password-locking \citep{vanderweij2025}, in-context scheming goals \citep{meinke2024scheming}, and fine-tuned sandbagging, were not tested.

The cyclic shift design tests absolute positional effects only. It cannot distinguish position-tracking from token-level letter priors. Because letter labels and displayed positions are confounded, the present design cannot determine whether the attractor is driven by option-label priors, serial position, or their interaction. Follow-up experiments with label remapping (replacing A--J with neutral tokens) or full random shuffle would help disambiguate.

Finally, the dataset is MMLU-Pro with 10 options per item. Results may not generalise to 4-option or 5-option benchmarks where the positional distribution space is smaller and biases may manifest differently.

\section{Conclusion}

Under sandbagging instruction, Llama-3-8B enters a distributional position attractor that is highly stable under complete option-content rotation. The pre-registered item-level diagnostic was ambiguous, but pre-specified aggregate distributional evidence (JS divergence $= 0.027$, Pearson $r = 0.9994$, identical modal positions, accuracy spiking to 72\% at the preferred position) provides strong evidence for a position-indexed selection policy that is highly stable and largely content-invariant at the aggregate level. The sandbagging instruction induces distributional selection bias rather than content-aware answer avoidance.

Response-position entropy is a promising candidate detection signal for this sandbagging mode. It is black-box, cheap, and orthogonal to accuracy. Future work should test whether this signal generalises across model families, scales, and sandbagging induction methods, and whether it persists at frontier scale where content-aware sandbagging strategies may emerge. Operational deployment as a detector requires baseline characterisation under diverse non-sandbagging prompts and adversarial robustness testing.

\subsubsection*{Broader Impact Statement}
This work contributes to sandbagging detection, which is relevant to AI safety and governance. If language models can suppress demonstrated capability on evaluations, safety-critical assessments may underestimate risk. The proposed detection signal is limited in scope and should not be used as a sole indicator of sandbagging in deployment settings without further validation.

\bibliography{references}
\bibliographystyle{tmlr}

\appendix
\section{Pre-registration}
\label{app:prereg}
The full pre-registration is available at \url{https://osf.io/efr6s}. No deviations from the pre-registered analysis plan occurred. The pre-registered primary diagnostic returned \emph{ambiguous}; the pre-registered stochastic robustness check returned \emph{mechanism unclear}. All aggregate distributional analyses (Pearson $r$, JS divergence, TVD, accuracy by position) were pre-specified as supporting analyses.

\end{document}